\documentclass[10pt,twocolumn,letterpaper]{article}

\usepackage{algorithm}
\usepackage{algpseudocode}
\usepackage[normalem]{ulem}

\newcommand{\ecolor}[1]{\textcolor{black}{#1}}

\usepackage[pagenumbers]{wacv} 

\definecolor{wacvblue}{rgb}{0.21,0.49,0.74}
\usepackage[pagebackref,breaklinks,colorlinks,allcolors=wacvblue]{hyperref}

\def\wacvPaperID{2570} 
\def\confName{WACV}
\def\confYear{2027}

\title{UI-VISA: U-Net Initialized Vascular Image Segmentation Architecture}

\author{Asees Kaur \quad Suzanne S. Sindi \quad Erica M. Rutter\\
University of California, Merced\\
{\tt\small akaur101@ucmerced.edu, ssindi@ucmerced.edu, erutter2@ucmerced.edu,}
}

\begin{document}
\maketitle
\begin{abstract}
Accurate segmentation of vascular structures in digital subtraction angiography (DSA) images remains challenging due to the thin, elongated, and branching nature of blood vessels. Pixel-wise deep learning approaches such as U-Net achieve strong general-purpose segmentation performance but often produce fragmented or discontinuous predictions in fine vascular regions, since they do not explicitly enforce structural connectivity. Region growing algorithms preserve spatial context and topological continuity, but are highly sensitive to seed point initialization and can be computationally expensive. We propose UI-VISA (U-Net Initialized Vascular Image Segmentation Architecture), a hybrid pipeline that combines the complementary strengths of both approaches. UI-VISA uses U-Net's foreground predictions as informed seed points for a CNN-guided region growing algorithm, which then iteratively refines the segmentation by enforcing local connectivity and recovering fine vessel details that U-Net alone tends to miss or over-predict. We evaluate UI-VISA against standalone U-Net and a prior region-growing-based method (VISA) using 5-fold cross-validation on 26 DSA images. UI-VISA achieves the highest mean Dice and clDice scores across folds, and a paired Wilcoxon signed-rank test shows the improvement in clDice is statistically significant ($p=0.023$), consistent with the method's design goal of preserving vascular connectivity, while the improvement in Dice does not reach significance ($p=0.104$).
\end{abstract}    
\section{Introduction}
\label{sec:intro}


With advancements in medical imaging techniques such as X-rays, ultrasound, computed tomography (CT), and magnetic resonance (MR) scans, computer-aided image segmentation can extract specific regions of interest (RoIs) for further evaluation, such as organ identification and tumor detection \cite{GUO2019229}. 
Biomedical image segmentation, which involves partitioning images into meaningful segments, is particularly significant in both research and clinical settings \cite{seo2020machine,papanastasiou2024attention}. By segmenting medical images, clinicians can identify and isolate regions of interest, such as organs or lesions, from complex CT and MRI scans \cite{Wang_2022}.

\begin{figure}[hbt!]
\begin{subfigure}{.5\columnwidth}
  \centering
  \includegraphics[width=\columnwidth]{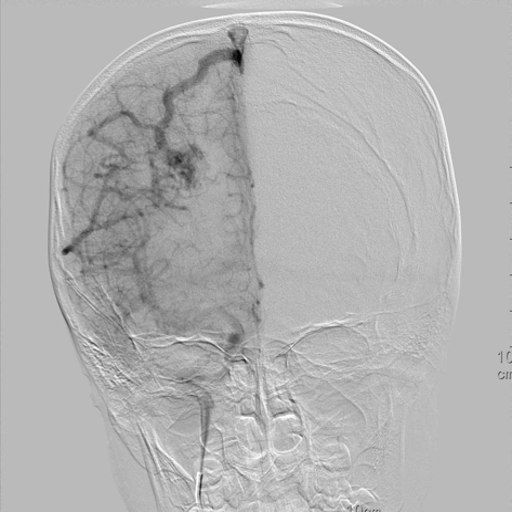}
  \caption{DSA Image}
\end{subfigure}%
\begin{subfigure}{.5\columnwidth}
  \centering
  \includegraphics[width=\columnwidth]{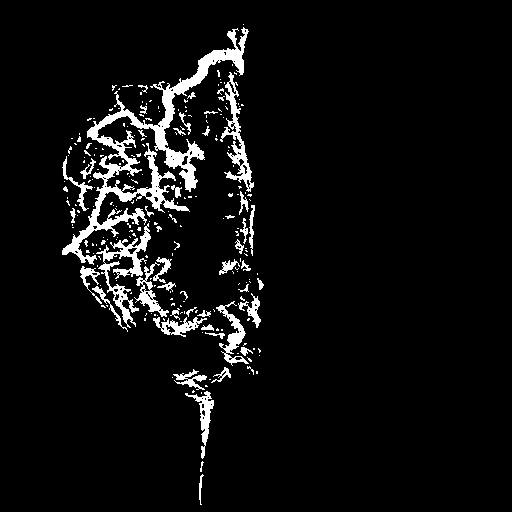}
  \caption{DSA Mask}
\end{subfigure}
\caption{Left (a) is an example of a typical DSA Image, acquired during angiography, and on the right (b) is its corresponding manually traced mask we use as ground truth.} 
\label{fig:data}
\end{figure}

Cerebrovascular diseases (CVDs) are conditions that slow blood flow to the brain, encompassing aneurysms, strokes, arteriovenous malformations, and arteriovenous fistulas, and \ecolor{are} one of the leading causes of death today \cite{zhang2020neural}. While CT and MR imaging techniques are used to detect strokes \cite{Nour2021StrokeImaging}, digital subtraction angiography (DSA) is the standard-of-care imaging method for diagnosing and guiding treatment \cite{Bhaskar2021DSA}, due to it's ability to clearly visualize blood vessels.  This extraction allows clinicians to make objective, data-driven decisions regarding vascular health. However, as illustrated in \Cref{fig:data}\ecolor{,} manually tracing vessel boundaries is tedious, error prone, and highly susceptible to inter- and intra-observer variability. 

To address these inconsistencies, machine learning methods---specifically convolutional neural networks (CNNs)---have become the standard for biomedical image segmentation tasks. Traditional techniques such as edge detection and thresholding have largely been surpassed by deep learning, which excels at hierarchical feature representation and produces strong segmentation results even under image noise, blur, or low contrast \cite{yao2023cnn,Wang_2022,yu2019techniques}.
Architectures such as U-Net \cite{ronneberger2015unet} are particularly effective in this domain, using an encoder-decoder structure and skip connections to learn localized features while retaining spatial information lost during downsampling. However, because U-Net processes the entire image or tile at once, it struggles with the fine, elongated, and branching structures characteristic of vascular imagery \cite{erutter}, often producing patchy probability maps with discontinuous vessel segments and isolated artifacts \cite{kimura2020virtual,ibtehaz2020multiresunet}. In addition to these architectural issues, medical image analysis often faces a shortage of labeled training data.

Region growing algorithms (RGAs) offer a complementary approach. Unlike U-Net, which classifies each pixel independently based on learned features, region growing makes each pixel's inclusion explicitly conditional on connectivity to the growing region, enforcing the topological continuity expected of vascular structures. Recent work proposed VISA (Vascular Image Segmentation Architecture), a two-step pipeline of a CNN-based region growing algorithm \cite{erutter, lagergren2023few, kaur2026thesis} that showed success in segmenting veinous structures. However, region growing is sensitive to the choice of initial seed points, as poor initialization can cause the algorithm to miss vessel branches or grow into irrelevant background regions \cite{Ma2020}.

To address these limitations, we propose UI-VISA (U-Net Initialized Vascular Image Segmentation Architecture), a hybrid pipeline that combines the strengths of both approaches. The modified U-Net first performs an initial segmentation and the resulting foreground predictions then serve as seed points for the RGA. UI-VISA refines the initial segmentation by enforcing local connectivity and recovering fine structural details that U-Net may have missed or incorrectly predicted, while also reducing unnecessary computation and eliminating seed bias. Our contributions with UI-VISA are three-fold:
\begin{enumerate}
    \item UI-VISA replaces random seeding \ecolor{in a CNN-based region growing algorithm} with smarter, informed seed selection guided by U-Net predictions. \ecolor{This effectively combines the strengths of U-Net and region growing algorithms.} 
    \item UI-VISA enforces local connectivity by exploiting region growing. \ecolor{Since the region-growing algorithm is applied \textit{after} the initial U-Net segmentation, we are able to correct patchy or non-contiguous segmentations.}
    \item UI-VISA achieves statistically significant \ecolor{improvement in} clDice scores compared to U-Net alone, demonstrating better preservation of vascular topology.
\end{enumerate}

\begin{figure*}[t]
  \centering
   \includegraphics[width=0.8\linewidth]{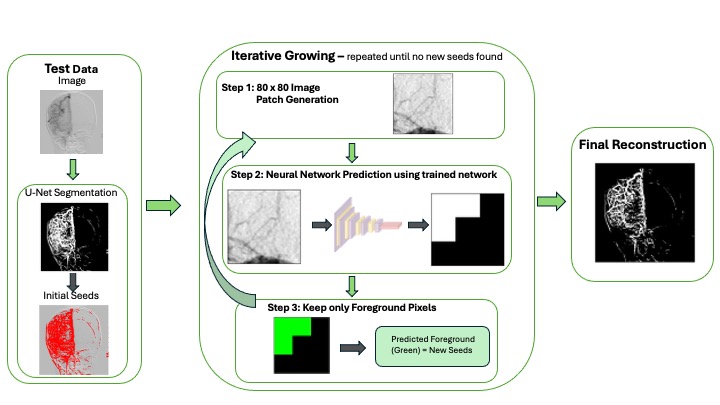}

   \caption{Proposed UI-VISA framework for vessel segmentation. The left panel shows a test image segmented by U-Net, with predicted foreground pixels (red) used as initial seeds. The middle panel illustrates the iterative seed expansion pipeline: $80 \times 80$ image patches centered on current seeds are fed into a trained CNN to predict $3 \times 3$ mask patches, with newly predicted foreground pixels added as seeds and the process repeated until convergence. The right panel shows the final reconstructed vessel segmentation.}
   \label{fig:unet_rgc_pipeline}
\end{figure*}

\noindent \ecolor{\Cref{fig:unet_rgc_pipeline} displays the pipeline of UI-VISA, in which a trained U-Net provides initial seeds (left) to a RGA (middle) to produce a final reconstruction (right).}

\section{Related Work}

\textcolor{black}{Our approach draws on two lines of prior work: automated segmentation methods for DSA imagery, and region growing algorithms for preserving vascular connectivity. We briefly describe both before presenting UI-VISA.}

\noindent
\subsection{Digital Subtraction Angiography Image Segmentation.} 

\textcolor{black}{Digital subtraction angiography (DSA) provides dynamic imaging of cerebral vessels with high spatial and temporal resolution and is considered the gold standard for diagnosing cerebrovascular diseases (CVDs) \cite{10884618, Heiss2001Imaging}. Extracting vessel structure from DSA images for further analysis has traditionally relied on manual tracing, which motivated the development of automated segmentation methods \cite{10884618}.} 

\textcolor{black}{Non-deep-learning-based approaches to DSA vessel segmentation use hand-crafted feature enhancement rather than learned representations. One method uses a multi-scale Hessian matrix approach that enhances vessel edges, filters out vessel-like noise, and fuses multiple frames from a DSA sequence to better visualize the overall vascular structure \cite{LUO2023100603}. However, the authors note that this multi-scale enhancement itself introduces blurring at vessel edges \cite{LUO2023100603}.}

\textcolor{black}{To reduce the dependence on manual and parameter-based methods, more recent work uses deep learning based approaches for vascular segmentation. CNNs are the most widely used, with U-Net \cite{ronneberger2015unet} being the most common architecture applied to DSA vessel segmentation \cite{zhang2020neural}.
A U-Net has also been pretrained on large amounts of unlabeled angiograms using an image-to-image translation framework on unannotated angiograms and then fine-tuned on a small set of labeled images to perform vessel segmentation \cite{Zeng2024Pretrained}. Furthermore, attention mechanisms, which allow the network to focus on most relevant spatial regions, have also been incorporated into U-Net to better capture the scale variation and fine vessel detail characteristic of cerebral DSA images \cite{Cui2023Spatial}. However, none of the above methods explicitly account for connectivity of the vessels, often resulting in fragmented or discontinuous segmentations.} 

\subsection{Region Growing Algorithms.}

\textcolor{black}{It has been recognized that classifying pixels independently - as many segmentation methods do - can yield predictions that are locally plausible but globally inconsistent with underlying structures, particularly through producing fragmented or disconnected regions that fail to respect contiguity of physical and biological objects \cite{haralick1985image}. Region growing algorithms (RGAs) were developed to address this limitation. Rather than treat each pixel independently, a pixel's inclusion in a region is made explicitly conditional on it's similarity to, and connectivity with, a set of neighboring pixels. Specifically, RGAs consider an initial seed and expand iteratively to neighboring pixels until no more similar pixels remain \cite{erutter}. Because the connectivity constraint is enforced, region growing approaches are especially well suited to vascular segmentation where the underlying biology consists of thin and branching structures.}

\textcolor{black}{Non-deep learning RGA variants have been applied directly to vascular data in several contexts. In \cite{7160191}, RGA was applied to vessel segmentation in MR angiography. Other work used frequency-domain information to automatically identify seed points for region growing, rather than placing them manually, by extracting a vessel's orientation from its spectral characteristics \cite{Jiang2013Region}. In a related approach, \cite{Ma2020} proposed a coronary artery method that adapts the search area at each growth step to follow vessel geometry.}

\textcolor{black}{More recently, region growing techniques have been combined with CNNs to replace the handcrafted similarity criteria with learned features. One method uses CNNs to predict neighboring pixel intensities to decide which pixels need to be added to the growing region \cite{erutter}. Other work has instead combined a CNN with a graph neural network to explicitly model the connectivity between vessel segments during segmentation \cite{Shin_2019}.}

\section{Methods}

\subsection{Data}
The dataset consists of 26 DSA images \ecolor{of size 512 $\times$ 512 pixels} from distinct patients and their manually traced masks \cite{zhang2020neural}. \ecolor{Images are normalized by dividing pixel intensities by the maximum value in the image. }We used 5-fold cross-validation in which the 26 image-mask pairs were randomly split into 5 folds. Since 26 does not divide evenly by 5, one fold contains 6 images for testing while the remaining four folds each contain 5, resulting in 20 or 21 images for training per fold respectively. The training set was further divided into 80\% training data and 20\% validation set used to select the best model weights and determine the threshold.  



\subsection{\ecolor{Modified }U-Net}
\label{sec:UNet}
To establish a baseline \ecolor{and compare with previous results \cite{zhang2020neural}, we trained} a modified U-Net architecture. Given the limited size of the dataset (26 DSA images of $512 \times 512$ pixels), patches of size $128 \times 128$ pixels were extracted as network inputs, following \cite{zhang2020neural}, with 200 patches randomly sampled per image to increase the effective training set size. Validation patches were extracted identically but without augmentation. Training patches were augmented via random $90^\circ$ rotations, horizontal/vertical flips, and random scaling (factor sampled uniformly from $[0.8, 1.2]$), each applied independently with probability 0.5 and identically to both image and mask; patches were re-cropped or reflection padded to maintain the fixed $128 \times 128$ input size. 
\begin{figure}[h]
    \includegraphics[width=\columnwidth]{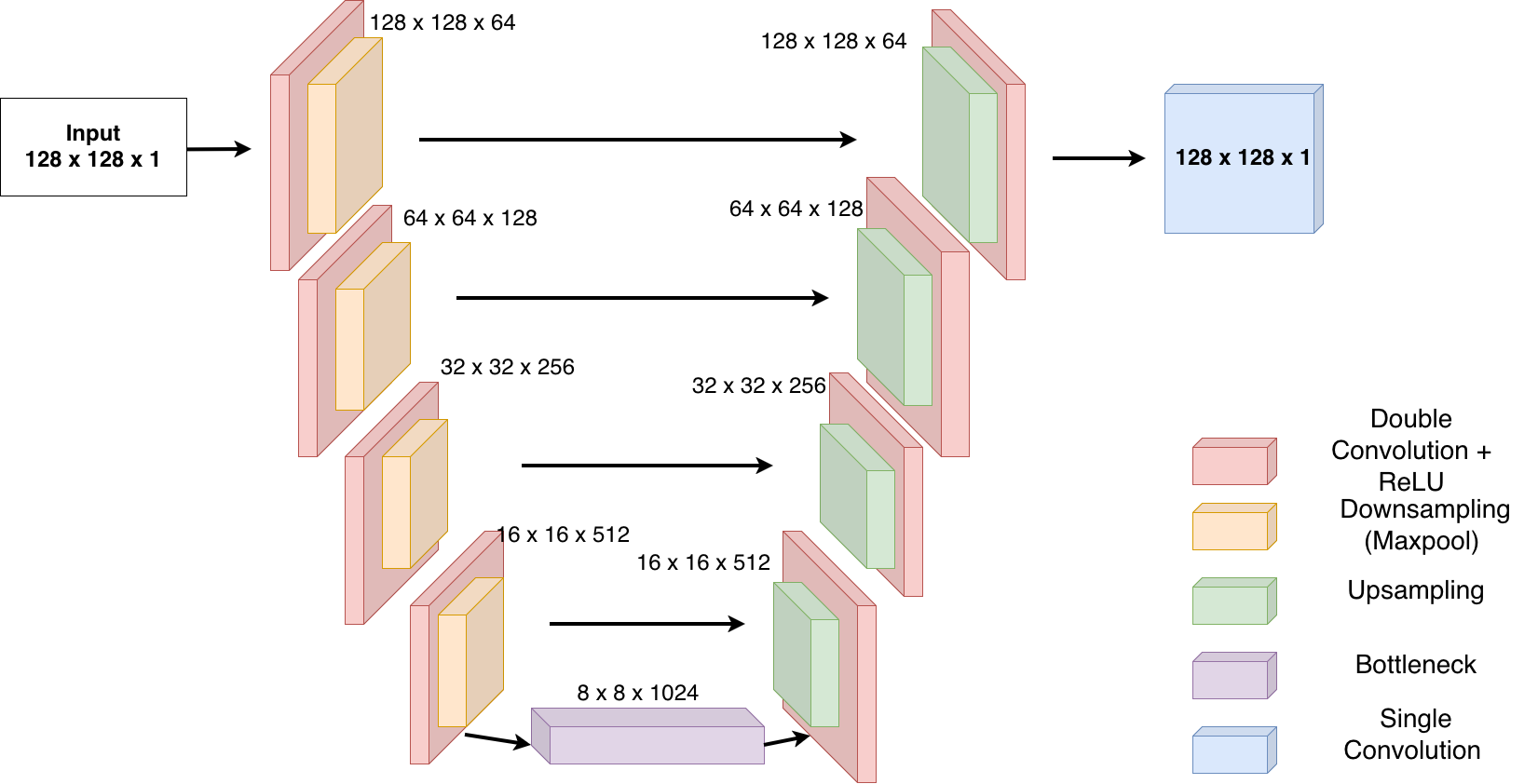}
   \caption{Modified U-Net Architecture. The encoder (left) consists of four double convolution blocks with ReLU activations, each followed by max pooling, progressively reducing the spatial dimensions from $128 \times 128$ to $8 \times 8$ while increasing the feature channels from 64 to 512. A bottleneck block at the deepest level expands the channels to 1024. The decoder (right) mirrors the encoder, restoring spatial resolution through upsampling and concatenating encoder feature maps via skip connections (horizontal arrows). A final $1 \times 1$ convolution produces the output segmentation mask of size $128 \times 128 \times 1$. }
    \label{fig:unet_cnn}
\end{figure}

The modified U-Net architecture, illustrated in \Cref{fig:unet_cnn}, follows an encoder-decoder design with skip connections. The encoder consists of four double $3\times3$ convolutional blocks (ReLU, dropout 0.2) with channels progressing from 64 to 512, followed by a 1024-channel bottleneck; the decoder mirrors this structure, ending in a $1\times1$ convolution producing a single-channel vessel probability map. The model was trained using SGD (momentum 0.9, initial learning rate 0.01, batch size 32) with binary cross-entropy loss on raw logits. Early stopping (patience 25 epochs, monitored on validation loss) was used, retaining the checkpoint with lowest validation loss as the final model for each fold.

\subsection{VISA: Vascular Image Segmentation Architecture}
\label{sec:visa}

\ecolor{VISA is a two-stage pipeline for CNN-based region rowing algorithm. In Stage 1, a patch-based CNN is trained on a small $80 \times 80 $ pixel image patch to predict foreground locations in a $3\times 3$ pixel mask. In Stage 2, a RGA uses the trained CNN from stage 1 to iteratively grow foreground pixels. Below, we describe each stage.}\\

\noindent
\ecolor{\textbf{Stage 1: CNN Training.}} 
\ecolor{To generate the patches for training, images and masks were zero-padded by 40 pixels on all sides (yielding $592 \times 592$ images) to ensure that regions centered near the original image boundaries remain fully defined. For each pixel $(i, j)$ in the original image, we extract an $80 \times 80$ patch from the padded image and a corresponding $3 \times 3$ patch from the padded mask, both centered at $(i+40, j+40)$. This yields 262,144 image–label pairs per image, one for every pixel}. To increase training diversity, we applied random $90^\circ$ rotations and horizontal/vertical flips to each training image and mask, each with probability 0.5. \textcolor{black}{Each transformation that was applied produced an additional image-mask pair, meaning each original training image could yield up to four versions (the original plus up to three augmented copies).} 

\begin{figure}[hbtp]
\centering
  \includegraphics[width=0.8\columnwidth]{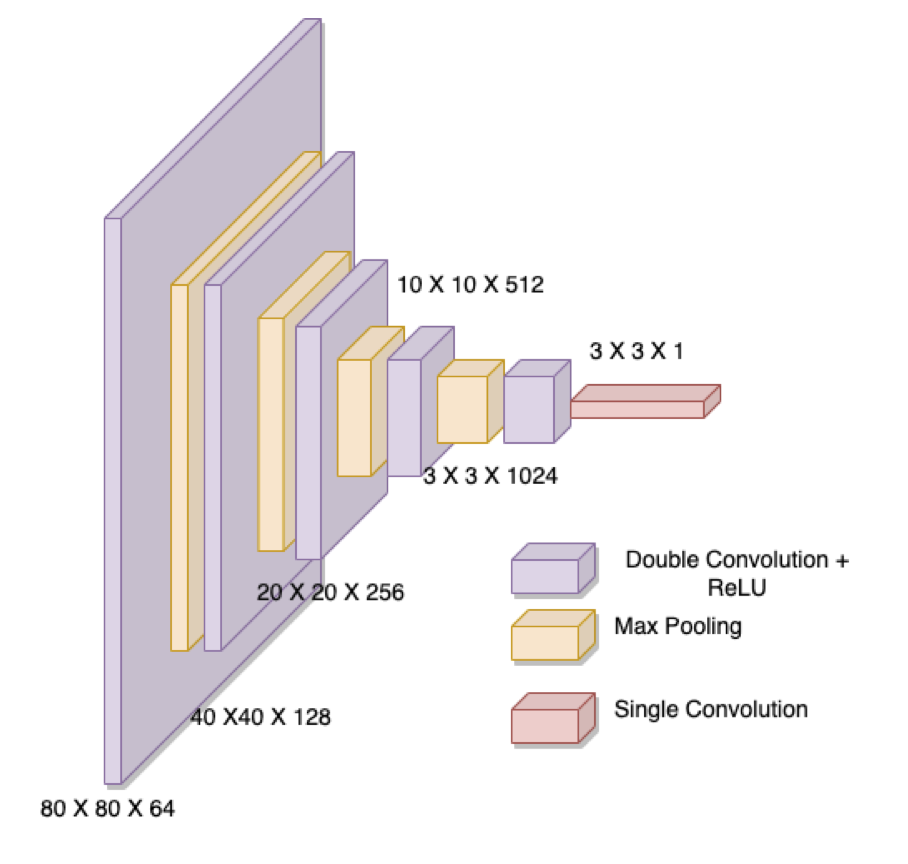}
  \caption{VISA Network Architecture: Convolutional neural network (CNN) architecture consisting of layers with double convolution and ReLU (purple), max pooling (yellow), and a final single convolution (red). The network takes an $80 \times 80 \times 1$ grayscale patch as input. The dimensions labeled on each layer reflect the output of that layer, starting from $80 \times 80 \times 64$ after the first double convolution and ending at $3 \times 3 \times 1$.}

  \label{fig:architecture}
\end{figure}

\textcolor{black}{Because vessel pixels (foreground) are vastly outnumbered by background pixels, we balanced the training data prior to training. Specifically, for each $3 \times 3$ mask patch, we computed the sum of its binary entries (0 = full background, 1 = full foreground), producing a value from 0 to 9. In each image, we identified the class with the minimum number of samples and then subsampled the remaining classes with replacement to ensure balanced representation.}

VISA uses a 25-layer CNN resembling the encoder portion of U-Net \cite{ronneberger2015unet}: five double-convolutional blocks (ReLU-activated), each followed by max pooling, and a final $1 \times 1$ convolution producing a $3 \times 3 \times 1$ output representing foreground/background probability. The architecture is shown in \Cref{fig:architecture}. \textcolor{black}{During training, the model was optimized using the Adam optimizer \cite{kingma2015adam} and the binary cross entropy loss function. Training was conducted on a GPU and the process spanned 50 epochs, with a batch size of 50 image-label pairs for the training data. To increase the diversity of training data seen by the model at each epoch, the balanced patch extraction procedure was re-run at the start of every epoch. This means a different random subset of patches was sampled from each category for each epoch, rather than reusing the same fixed set of patches throughout training.}\ecolor{ To make a binary prediction, the probability of foreground was selected per fold based on validation set experiments, ranging from $0.4$ to $0.6$, where pixels with a predicted probability greater than the threshold are classified as foreground.}\\ 

\begin{figure}[htbp]
  \centering
  \begin{subfigure}[b]{0.495\columnwidth}
    \includegraphics[width=\linewidth]{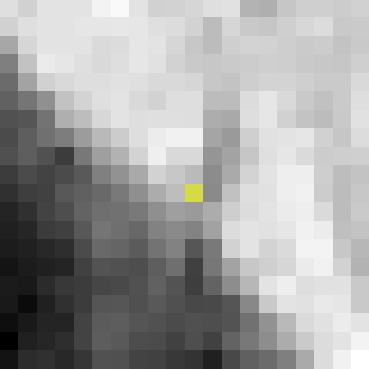}
    \caption{Single initial foreground seed pixel in yellow}
    \label{fig:rgc1}
  \end{subfigure}
  \begin{subfigure}[b]{0.495\columnwidth}
    \includegraphics[width=\linewidth]{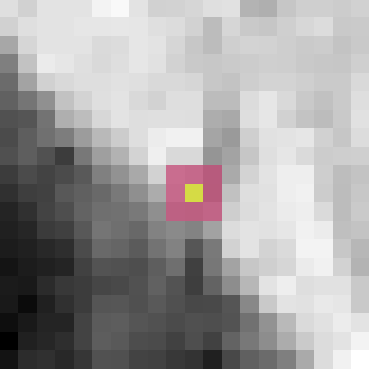}
    \caption{Candidate pixels that will be predicted by the trained network in red}
    \label{fig:rgc2}
  \end{subfigure}
  
  \begin{subfigure}[b]{0.495\columnwidth}
    \includegraphics[width=\linewidth]{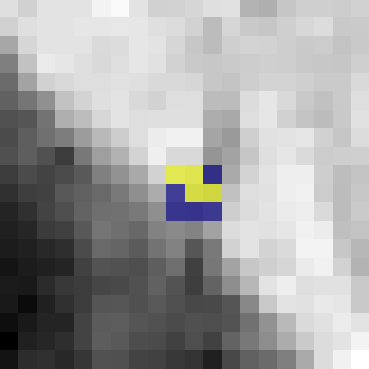}
    \caption{Foreground (yellow) and background (purple) predicted pixels}
    \label{fig:rgc3}
  \end{subfigure}
  \begin{subfigure}[b]{0.495\columnwidth}
    \includegraphics[width=\linewidth]{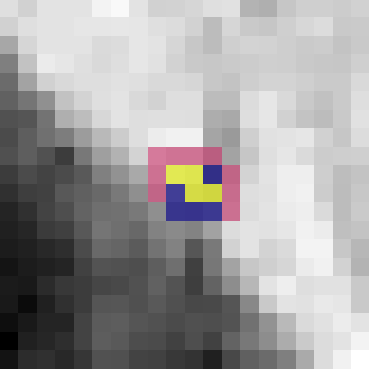}
    \caption{Next iteration of candidate pixels (red) to be predicted by network}
    \label{fig:rgc4}
  \end{subfigure}
  \caption{Illustration of Stage 2 of VISA. A single initial seed pixel (yellow) is shown in a. Panel b identifies the next candidate pixels (red) that will be evaluated by the trained network. Panel c shows the predicted classification of pixels into foreground (yellow) and background (purple). Panel d highlights the next set of candidate pixels (red) for the subsequent iteration of the network evaluation.}
  \label{fig:rgc_method}

\end{figure}

\noindent
\ecolor{\textbf{Stage 2: Region Growing Algorithm.} Once the CNN is trained, the RGA is used to generate a contiguous vessel segmentation efficiently without processing all pixels in the image. Starting from a set of initial seed pixels (denoted by the yellow pixel in \Cref{fig:rgc_method}a), the algorithm extracts an 80$\times$80 image patch around each seed, passes it through the trained network from Stage 1, and obtains a 3$\times$3 prediction of foreground/background centered around that seed (\Cref{fig:rgc_method}b,c). Pixels predicted as foreground are added to the segmented region and the algorithm repeats the process on their neighbors -- extracting new patches, predicting, and expanding only in the direction of foreground pixels (\Cref{fig:rgc_method}d). The algorithm terminates when no new foreground pixels are found. In practice, we start with 500 initially selected random pixels in our $512 \times 512$ image to serve as our seeds. } 

\subsection{Proposed method: UI-VISA}

\ecolor{UI-VISA attempts to mitigate the stochastic nature of VISA predictions by replacing the random selection of seed points. Stage 1 of UI-VISA and VISA are identical. The only difference in Stage 2 of UI-VISA is that we replace the 500 randomly selected initialization pixels with the predicted foreground pixels of U-Net. The subsequent RGA is identical to VISA. }As illustrated in \Cref{fig:unet_rgc_pipeline}, the UI-VISA testing pipeline begins with the \ecolor{modified} U-Net \ecolor{(see \Cref{sec:UNet})} producing a full segmentation of the test image, as shown in the left panel of the figure. The U-Net output is thresholded at 0.5 \ecolor{to produce a segmentation mask.} The resulting foreground pixels, shown in red at bottom of the left panel, are then used as the initial seeds for the RGA, the second \ecolor{stage} of VISA. Since each pixel may be predicted multiple times due to the overlapping nature of the patches, the predicted probabilities are averaged across all overlapping predictions before the final threshold is applied. \ecolor{See \Cref{alg:hybrid} for full details.}

It is important to note that the final segmentation is produced entirely by the RGA, not U-Net. The U-Net predictions serve solely as an informed starting point, \ecolor{and }the RGA then refines this initialization by expanding around the seed pixels. \ecolor{This refinement} recovers fine vessel details and enforces local connectivity in regions where the \ecolor{modified} U-Net may have produced fragmented or incomplete predictions. This refinement step is critical for vascular structures, where connectivity and continuity are clinically important. \ecolor{In addition to adding foreground, the refinement can also eliminate false positive foreground pixels segmented by U-Net.}

\begin{algorithm}
\caption{UI-VISA Testing Algorithm}
\label{alg:hybrid}
\begin{algorithmic}[1]
\Require Trained U-Net $g$, trained VISA CNN $f$, test image $I$, threshold $\tau$
\Ensure Predicted segmentation mask $\hat{Y}$
\State Normalize $I$ by dividing by its maximum pixel value
\State Pad $I$ with 40 zeros on all sides to obtain $I_{\text{pad}}$
\State Initialize $\hat{Y}(i,j) \leftarrow 0$ for all pixels $(i,j)$
\State Initialize pixel score dictionary $\mathcal{P} \leftarrow \{\}$
\State Initialize visited set $S \leftarrow \emptyset$
\State Obtain U-Net foreground predictions: $\mathcal{C}_0 \leftarrow \{(i,j) : g(I)(i,j) \geq 0.5\}$
\State $t \leftarrow 0$
\While{$\mathcal{C}_t \neq \emptyset$}
    \ForAll{$(i,j) \in \mathcal{C}_t$}
        \State Extract $80 \times 80$ patch $\mathcal{S}(i,j)$ from $I_{\text{pad}}$ centered at $(i,j)$
        \State Obtain $3 \times 3$ prediction mask $P \leftarrow f(\mathcal{S}(i,j))$
        \ForAll{$(\text{row}, \text{col}) \in P$}
            \State Append $P[\text{row}, \text{col}]$ to $\mathcal{P}[(i+\text{row}-1, j+\text{col}-1)]$
        \EndFor
        \State $S \leftarrow S \cup \{(i,j)\}$
    \EndFor
    \State $\mathcal{C}_{t+1} \leftarrow \{(i,j) \notin S : \mathrm{mean}(\mathcal{P}[(i,j)]) \geq \tau\}$
    \State $t \leftarrow t + 1$
\EndWhile
\State \textbf{Reconstruct mask:}
\ForAll{$(i,j) \in \mathcal{P}$}
    \If{$\mathrm{mean}(\mathcal{P}[(i,j)]) \geq \tau$}
        \State $\hat{Y}(i,j) \leftarrow 1$
    \EndIf
\EndFor
\State Remove padding by cropping rows and columns by 40 of $\hat{Y}$ to recover the original $512 \times 512$ mask
\State \Return $\hat{Y}$
\end{algorithmic}
\end{algorithm}

\subsection{Evaluation Metrics}
\ecolor{To quantify the accuracy of segmentation, we use both Dice score and clDice score. }
While the Dice score measures the overall overlap between the predicted segmentation and the ground truth, it does not explicitly account for the connectivity or topology of thin, elongated structures such as blood vessels. As a result, segmentations with similar Dice scores may differ substantially in terms of vascular continuity, with breaks or disconnections that are clinically relevant. To address this limitation, we also evaluate segmentation performance using centerline Dice (clDice) \cite{shit2021cldice}, a metric specifically designed for vascular structures. clDice assesses how well the centerlines of the predicted and ground truth segmentations align with each other, thereby emphasizing connectivity and structural correctness.
It is computed by first extracting the centerlines (skeletons) of both the predicted segmentation and the ground truth mask. The clDice score is then defined as:
\begin{equation}\text{clDice(skeleton)} = \frac{2 \times | \text{Skeleton}_{\text{seg}} \cap \text{Skeleton}_{\text{gt}} |}{|\text{Skeleton}_{\text{seg}}| + |\text{Skeleton}_{\text{gt}}|}.
\end{equation}
A higher clDice value indicates better preservation of vessel continuity and topology. Here, $\text{Skeleton}_{\text{seg}}$ represents the skeleton of the segmented image, and $\text{Skeleton}_{\text{gt}}$ represents the skeleton of the ground truth mask. 

\section{Results}

\begin{figure*}[t]
\centering
\setlength{\tabcolsep}{2pt}
\begin{tabular}{cccc}

\textbf{Original} & \textbf{VISA} & \textbf{U-Net} & \textbf{UI-VISA} \\[4pt]

\includegraphics[width=0.23\textwidth]{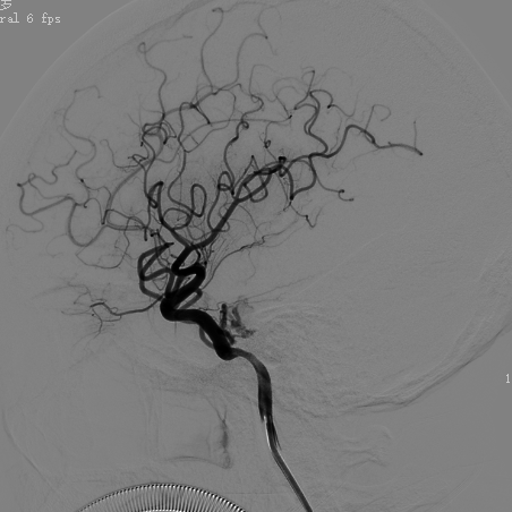} &
\includegraphics[width=0.23\textwidth]{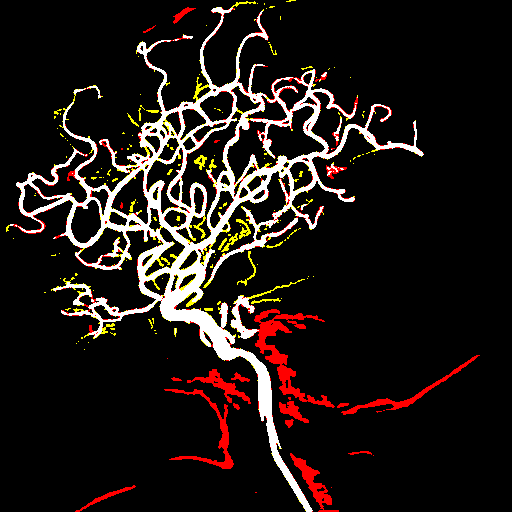} &
\includegraphics[width=0.23\textwidth]{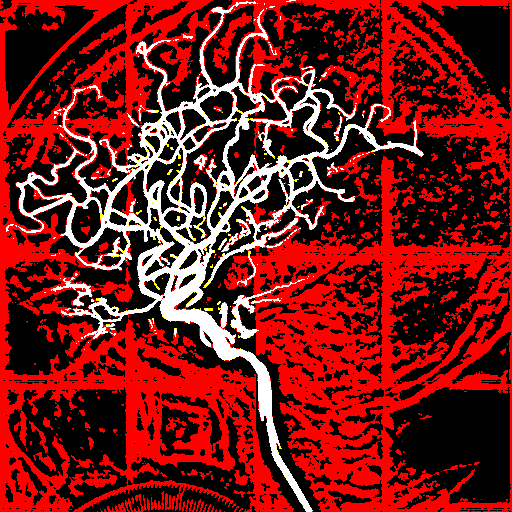} &
\includegraphics[width=0.23\textwidth]{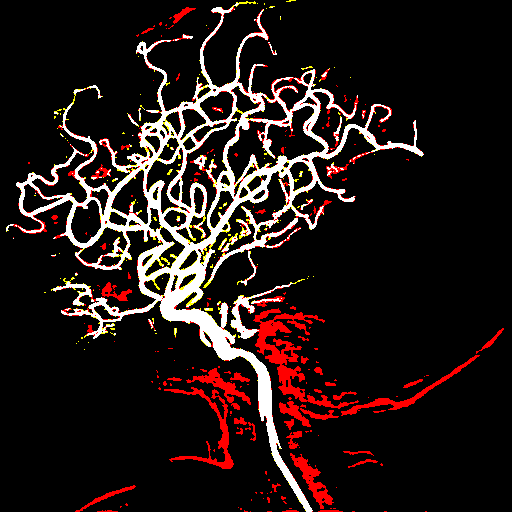} \\
{\scriptsize \textcolor{white}{\shortstack{Dice: 0.7925 \\ clDice: 0.8076}}} &
{\scriptsize \shortstack{Dice: 0.7925 \\ clDice: 0.8076}} &
{\scriptsize \shortstack{Dice: 0.3216 \\ clDice: 0.3409}} &
{\scriptsize \shortstack{Dice: 0.7694 \\ clDice: 0.7844}} \\[6pt]

\includegraphics[width=0.23\textwidth]{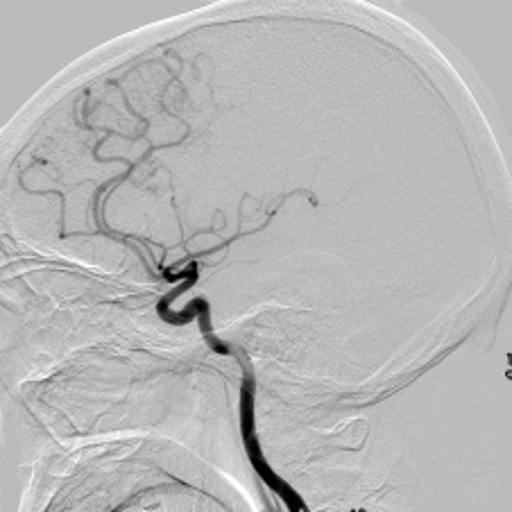} &
\includegraphics[width=0.23\textwidth]{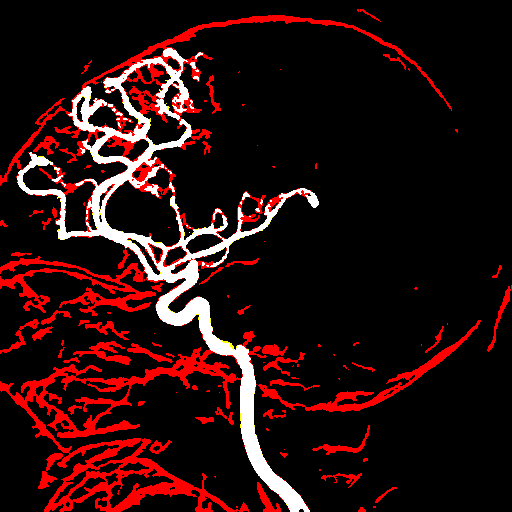} &
\includegraphics[width=0.23\textwidth]{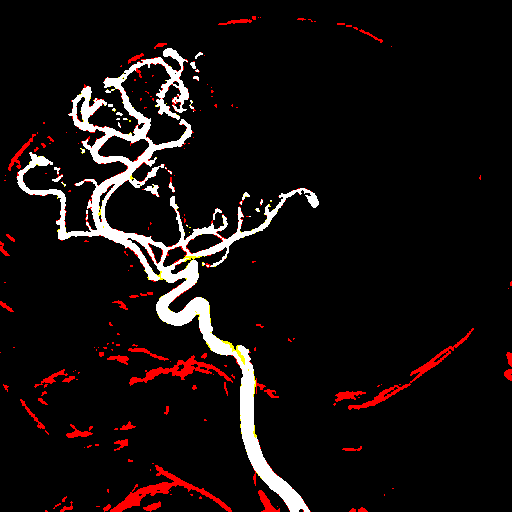} &
\includegraphics[width=0.23\textwidth]{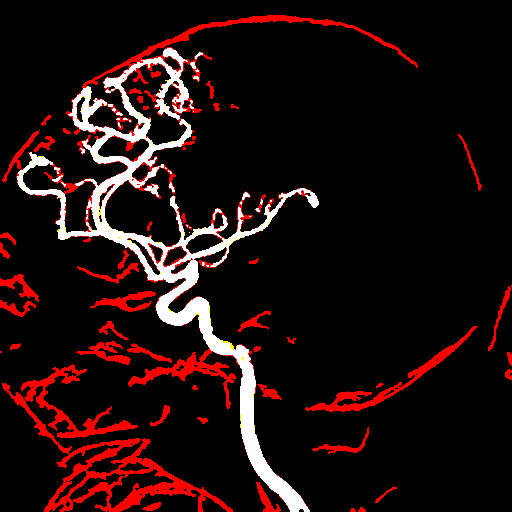} \\
{\scriptsize \textcolor{white}{\shortstack{Dice: X.XX \\ clDice: X.XX}}} &
{\scriptsize \shortstack{Dice: 0.5102 \\ clDice: 0.4354}} &
{\scriptsize \shortstack{Dice: 0.7920 \\ clDice: 0.7260}} &
{\scriptsize \shortstack{Dice: 0.5963 \\ clDice: 0.5555}} \\[6pt]

\includegraphics[width=0.23\textwidth]{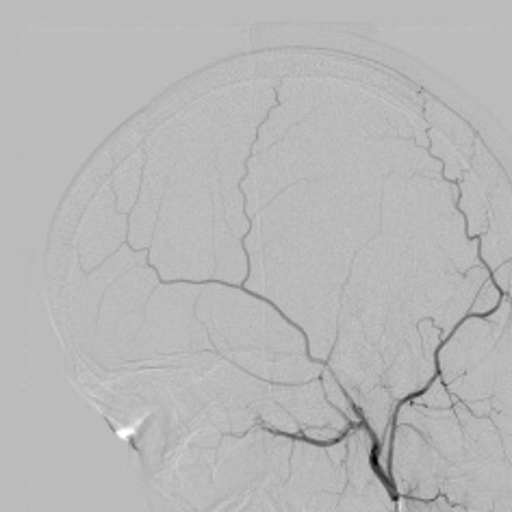} &
\includegraphics[width=0.23\textwidth]{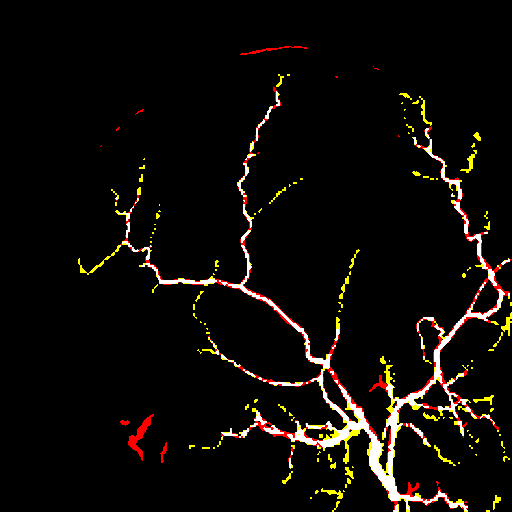} &
\includegraphics[width=0.23\textwidth]{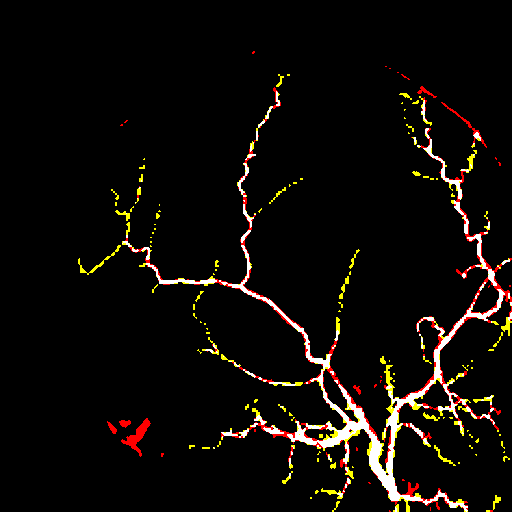} &
\includegraphics[width=0.23\textwidth]{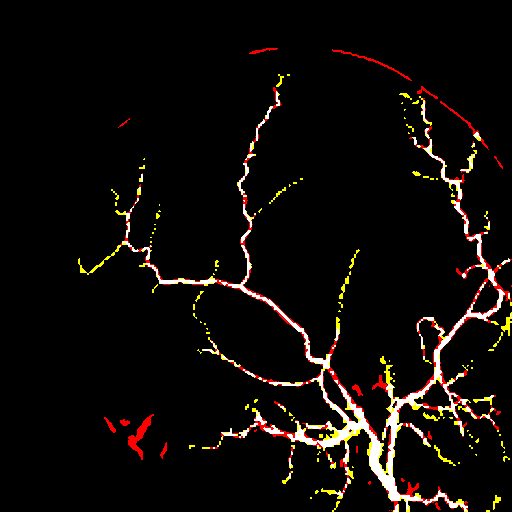} \\
{\scriptsize \textcolor{white}{\shortstack{Dice: X.XX \\ clDice: X.XX}}} &
{\scriptsize \shortstack{Dice: 0.6519 \\ clDice: 0.6927}} &
{\scriptsize \shortstack{Dice: 0.6577 \\ clDice: 0.6984}} &
{\scriptsize \shortstack{Dice: 0.6559 \\ clDice: 0.7003}} \\

\end{tabular}

\vspace{0.5em}
\includegraphics[width=1\textwidth]{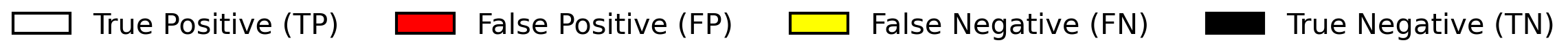}

\caption{Qualitative comparison of segmentation outputs from VISA, U-Net, and UI-VISA approaches across three test images. Dice and clDice scores are reported under each prediction. }
\label{fig:grid}
\end{figure*}

\Cref{fig:grid} presents a qualitative comparison of the three methods across three representative test images. The images use a color-coded scheme to visualize prediction accuracy: white represents true positives (correctly identified vessel pixels), black represents true negatives (correctly identified background), red represents false positives (background incorrectly classified as vessel), and yellow represents false negatives (vessel pixels incorrectly classified as background). \ecolor{Three different scenarios were selected to show the breadth of results.} In the first row, VISA \ecolor{and UI-VISA vastly outperform the modified U-Net, which produces a poor segmentation.} \ecolor{UI-VISA }recovers well with a Dice of $0.7694$, demonstrating that even when U-Net predictions are fragmented, the RGA refinement stage produces a coherent segmentation. In the second row, \ecolor{we present a case in which} U-Net achieves the best Dice of $0.7920$, while both VISA ($0.5102$) and \ecolor{UI-VISA}($0.5963$) struggle. 
\ecolor{It appears that UI-VISA introduces extra false positives by including the skull.} In the third row, all three methods perform similarly, 
suggesting that this particular image is equally challenging for all approaches. 

\begin{table*}[t]
\centering
\caption{Summary of Cross-Validation Results for VISA, U-Net, and UI-VISA Models}
\label{tab:cv_summary}
\resizebox{\textwidth}{!}{%
\begin{tabular}{c|cc|cc|cc}
\toprule
 & \multicolumn{2}{c|}{\textbf{VISA}} & \multicolumn{2}{c|}{\textbf{U-Net}} & \multicolumn{2}{c}{\textbf{UI-VISA}} \\
\cmidrule(lr){2-3} \cmidrule(lr){4-5} \cmidrule(lr){6-7}
\textbf{Fold} & \textbf{Dice} & \textbf{clDice} & \textbf{Dice} & \textbf{clDice} & \textbf{Dice} & \textbf{clDice} \\
\midrule
0 & $0.8126 \pm 0.0290$ & $0.7991 \pm 0.0597$ & $0.8165 \pm 0.0597$ & $0.7898 \pm 0.0796$ & $\mathbf{0.8227 \pm 0.0363}$ & $\mathbf{0.8084 \pm 0.0519}$ \\
1 & $0.8133 \pm 0.0431$ & $0.7909 \pm 0.0494$ & $0.8106 \pm 0.0507$ & $0.7898 \pm 0.0612$ & $\mathbf{0.8225 \pm 0.0438}$ & $\mathbf{0.8074 \pm 0.0568}$ \\
2 & $0.7433 \pm 0.0406$ & $0.7178 \pm 0.0569$ & $0.6905 \pm 0.1910$ & $0.6655 \pm 0.1754$ & $\mathbf{0.7695 \pm 0.0471}$ & $\mathbf{0.7600 \pm 0.0567}$ \\
3 & $0.7795 \pm 0.0692$ & $0.7714 \pm 0.0466$ & $0.7747 \pm 0.0687$ & $0.7596 \pm 0.0600$ & $\mathbf{0.7829 \pm 0.0724}$ & $\mathbf{0.7823 \pm 0.0522}$ \\
4 & $0.7585 \pm 0.1347$ & $0.7358 \pm 0.1604$ & $\mathbf{0.7942 \pm 0.0530}$ & $\mathbf{0.7634 \pm 0.0586}$ & $0.7759 \pm 0.1041$ & $0.7587 \pm 0.1169$ \\
\hline
All images & $0.7810 \pm 0.0725 $ & $ 0.7610 \pm 0.0887 $
& $ 0.7788 \pm 0.1110 $ & $ 0.7550 \pm 0.1095 $
& $ \mathbf{0.7958 \pm 0.0702} $ & $ \mathbf{0.7843 \pm 0.0756}$ \\
\bottomrule
\end{tabular}}
\end{table*}

\Cref{tab:cv_summary} summarizes the cross-validation results for the UI-VISA, U-Net, and VISA models across all five folds\ecolor{, with highest values highlighted in bold}.  UI-VISA achieves the highest Dice and clDice scores in four out of five folds, demonstrating that replacing random seed initialization with informed U-Net predictions consistently improves segmentation performance \ecolor{over the random-seed initialization in VISA}. Across folds 0--3, the hybrid model outperforms both standalone methods in both metrics, with Dice scores ranging from $0.7695$ to $0.8227$ and clDice scores ranging from $0.7600$ to $0.8084$. The only exception is Fold 4, where U-Net achieves the highest Dice ($0.7942$) and clDice ($0.7634$), while \ecolor{UI-VISA} scores $0.7759$ and $0.7587$ respectively. \ecolor{Even in this challenging case, we note that UI-VISA still outperforms VISA, implying that the more stable seed initialization is an improvement.} \textcolor{black}{The last row of the table shows the average of the Dice and clDice scores over all images, since each image is contained in the test dataset exactly once. The results over all images indicate that UI-VISA is the highest-performing algorithm.} 

\begin{table}[h]
\centering
\caption{Testing Time per Fold for VISA, U-Net, and UI-VISA Models}
\label{tab:time_summary}
\begin{tabular}{c|c|c|c}
\toprule
\textbf{Fold} & \textbf{VISA (s)} & \textbf{U-Net (s)} & \textbf{UI-VISA (s)} \\
\midrule
0 & $3074.99$ & $38.60$ & $1043.67$ \\
1 & $2222.99$ & $32.70$ & $695.17$ \\
2 & $4420.55$ & $39.95$ & $1004.08$ \\
3 & $2436.47$ & $32.53$ & $592.30$ \\
4 & $9341.06$ & $33.26$ & $1811.65$ \\
\bottomrule
\end{tabular}
\end{table}

\Cref{tab:time_summary} reports the inference time per fold for each method. U-Net is by far the fastest, completing inference in under 40 seconds per fold, since it only requires a forward pass over extracted patches without any iterative expansion. However, speed alone does not reflect segmentation quality for vascular structures, where connectivity and continuity are biologically essential properties. As demonstrated in \Cref{tab:cv_summary}, U-Net consistently produces lower clDice scores than the hybrid approach across most folds, indicating that its pixel-wise predictions fail to preserve the topological structure of the vascular network, a critical limitation for clinical applications where vessel continuity directly informs diagnosis and treatment planning. \ecolor{While UI-VISA is significantly slower than U-Net, it is much faster than VISA, with improvement speeds ranging from 3-5 fold. We hypothesize that UI-VISA is quicker as it is only performing refinement instead of processing the entire image.}

\ecolor{While the previous results in \Cref{fig:grid} and \Cref{tab:cv_summary} suggest that UI-VISA outperforms the other methods, we aimed to quantify if the improved performance was statistically significant.} To validate the UI-VISA results, we performed a statistical hypothesis test comparing per-image performance against the U-Net baseline:
\begin{center}
   $\begin{aligned}
H_0&: \text{UI-VISA and U-Net perform equally,} \\
H_1&: \text{UI-VISA outperforms U-Net}.
\end{aligned}$
\end{center}
Because each of the 26 images appears in exactly one test fold, we obtained a single Dice and clDice score per image for both methods. For each metric, we computed the per-image difference between UI-VISA and U-Net and applied the Shapiro-Wilk test to assess whether these differences were normally distributed; \textcolor{black}{for both metrics the p-value was below a significance level of $\alpha = 0.001$ (Dice: $p = 3\times10^{-6}$; clDice: $p = 7\times10^{-6}$), so we used the Wilcoxon signed-rank test rather than a paired $t$-test.}

The Wilcoxon signed-rank test is a nonparametric test for paired data. It ranks the per-image differences between the two methods by size and checks whether they tend to fall on one side of zero, which would indicate that one method consistently outperforms the other \cite{wilcoxon1945}. 
For both Dice and clDice metrics, we computed the per-image difference between UI-VISA and U-Net across all 26 test images and used the one-sided Wilcoxon signed-rank test to evaluate our hypothesis that UI-VISA outperforms U-Net, at a significance level of $\alpha = 0.05$.

UI-VISA scores higher on 17 of the 26 images for Dice and 18 of 26 for clDice. Each of the differences $d$ was ranked by absolute value, and the ranks with positive sign were summed. Under the null hypothesis this sum would be close to the value expected if the signs were assigned at random. The p-value is the probability of observing a sum at least this large if the two methods performed equally. Using a significance level of $\alpha = 0.05$, the improvement in clDice was statistically significant (p = 0.023), while the improvement in Dice was not (p = 0.104). This is consistent with what UI-VISA is designed to do. clDice measures how well the vessel structure is preserved, while Dice measures pixel overlap, so an improvement in clDice suggests the RGA recovers connectivity that U-Net misses.
\section{Discussion and Future Work}

In this work, we presented UI-VISA (U-Net Initialized Vascular Image Segmentation Architecture), a hybrid pipeline that combines U-Net predictions with the region growing algorithm (RGA) of VISA to improve seed initialization and produce more accurate and connected vessel segmentations. The key distinction between the standalone VISA and UI-VISA lies in seed initialization. In standalone VISA, 500 seeds are selected randomly from the image, meaning they can fall anywhere in the image including background regions. This results in a large number of initial candidate pixels that must be processed and discarded before the algorithm converges on meaningful vessel structures, driving up computational cost. In UI-VISA, seeds are drawn entirely from U-Net foreground predictions, meaning every seed is already located within a predicted vessel region. This informed initialization eliminates unnecessary expansion into background regions and significantly reduces the total number of iterations required for convergence. As a result, UI-VISA is substantially faster than standalone VISA, ranging from approximately 592 to 1812 seconds per fold compared to 2223 to 9341 seconds for standalone VISA, a reduction of roughly 3 to 5 times. 

Taken together, these results demonstrate that UI-VISA gives the most favorable balance among the three methods: it is more accurate and better connected than U-Net alone, more computationally efficient than standalone VISA, and consistently preserves the vascular connectivity that is essential for reliable DSA image segmentation. The reduction in inference time relative to standalone VISA, combined with improved or comparable segmentation accuracy, supports UI-VISA as a practical and effective approach for DSA vessel segmentation. 

Despite the computational savings achieved through informed seed initialization, UI-VISA remains substantially slower than standalone U-Net inference. A promising direction for future work is to further reduce computational cost while preserving the connectivity advantages of the region growing approach. Rather than initializing seeds from all U-Net foreground predictions, a more targeted strategy would be to restrict seed selection to pixels lying on the boundary between predicted foreground and background regions in the U-Net output. Additionally, to avoid unnecessary computation, the RGA could be augmented with an early termination criterion: if the VISA CNN prediction at a given location agrees with the U-Net prediction, expansion in that direction is halted. This boundary-aware initialization strategy has the potential to significantly reduce the number of iterations required for convergence, bringing the computational cost of UI-VISA closer to that of standalone U-Net while retaining the topological correctness that makes region growing essential for vascular segmentation. 
\section{Acknowledgments}

 Part of this research was conducted using MERCED cluster (NSF-MRI, \#1429783) and Pinnacles (NSF MRI, \# 2019144) at the Cyberinfrastructure and Research Technologies (CIRT) at University of California, Merced.


\end{document}